\documentclass[10pt,twocolumn,aps,pre,superscriptaddress]{revtex4-2}
\usepackage[utf8]{inputenc}
\usepackage{amsmath,amsfonts,amssymb}
\usepackage{graphicx}
\usepackage{array}
\usepackage{bm}
\usepackage{quotes}
\usepackage{hyperref}
\usepackage{xcolor}
\usepackage{ulem}
\definecolor{goodblue}{RGB}{0, 91, 187}
\hypersetup{
  colorlinks=true,
  allcolors=goodblue,
  urlcolor=goodblue,
  citecolor=goodblue,
  pdfborder={0 0 0},
  breaklinks=true,
}

\DeclareMathOperator*{\argmin}{arg\,min}
\newcommand{\boldf}{\bm{f}}
\newcommand{\boldr}{\bm{r}}
\newcommand{\boldu}{\bm{u}}
\newcommand{\boldz}{\bm{z}}
\newcommand{\boldA}{\bm{A}}
\newcommand{\boldF}{\bm{F}}
\newcommand{\boldRW}{\bm{R}_{\text{w}}}
\newcommand{\boldV}{\bm{V}}
\newcommand{\boldVG}{\bm{V}_{\text G}}
\newcommand{\boldVD}{\bm{V}_{\text D}}
\newcommand{\Win}{\bm{W}_{\text{in}}}
\newcommand{\Wout}{\bm{W}_{\text{out}}}
\newcommand{\ER}{Erd\"os-R\'enyi }
\newcommand{\CG}{C_{\text{G}}}
\newcommand{\RG}{R_{\text{G}}}
\newcommand{\Vbias}{V_{\text{bias}}}
\newcommand{\Kp}{K_{\text{p}}}
\newcommand{\Vp}{V_{\text{p}}}
\newcommand{\VG}{V_{\text{G}}}
\newcommand{\VD}{V_{\text{D}}}
\newcommand{\IG}{I_{\text{G}}}
\newcommand{\ID}{I_{\text{D}}}
\newcommand{\Ich}{I_{\text{ch}}}
\begin{document}

\title{A theoretical framework for reservoir computing on networks of organic electrochemical transistors}

\author{Nicholas W. Landry}
\email{nicholas.landry@virginia.edu}
\affiliation{Department of Biology, University of Virginia, Charlottesville, Virginia, USA}
\affiliation{Vermont Complex Systems Center, University of Vermont, Burlington, Vermont, USA}
\affiliation{Department of Applied Mathematics, University of Colorado Boulder, Boulder, Colorado, USA}

\author{Beckett R. Hyde}
\email{beckett.hyde@colorado.edu}
\affiliation{Department of Applied Mathematics, University of Colorado Boulder, Boulder, Colorado, USA}

\author{Jake C. Perez}%
\email{jake.perez@colorado.edu}
\affiliation{Department of Electrical, Computer, and Energy Engineering, University of Colorado Boulder, Boulder, Colorado, USA}

\author{Sean E. Shaheen}%
\email{sean.shaheen@colorado.edu}
\affiliation{Department of Electrical, Computer, and Energy Engineering, University of Colorado Boulder, Boulder, Colorado, USA}
\affiliation{Department of Physics, University of Colorado Boulder, Boulder, Colorado, USA}
\affiliation{Renewable and Sustainable Energy Institute (RASEI), University of Colorado Boulder, Boulder, Colorado, USA}

\author{Juan G. Restrepo}%
\email{juanga@colorado.edu}
\affiliation{Department of Applied Mathematics, University of Colorado Boulder, Boulder, Colorado, USA}

\date{\today}

\begin{abstract}
Efficient and accurate prediction of physical systems is important even when the rules of those systems cannot be easily learned.
Reservoir computing, a type of recurrent neural network with fixed nonlinear units, is one such prediction method and is valued for its ease of training.
Organic electrochemical transistors (OECTs) are physical devices with nonlinear transient properties that can be used as the nonlinear units of a reservoir computer.
We present a theoretical framework for simulating reservoir computers using OECTs as the non-linear units as a test bed for designing physical reservoir computers.
We present a proof of concept demonstrating that such an implementation can accurately predict the Lorenz attractor with comparable performance to standard reservoir computer implementations.
We explore the effect of operating parameters and find that the prediction performance strongly depends on the pinch-off voltage of the OECTs.
\end{abstract}

\maketitle

\section{Introduction}

Reservoir computers are machine learning architectures where a high-dimensional, nonlinear dynamical system (the {\it reservoir}) is forced with a signal and a readout matrix is chosen to construct an output that performs a desired task on the input signal.
Reservoir computers (also referred to as RCs) are particularly suited to time-varying problems and have been successfully used for replicating and predicting  the evolution of dynamical systems \cite{pathak_using_2017}, for developing data-based control schemes \cite{canaday_model-free_2021}, for extrapolating dynamics to unobserved regions of parameter space \cite{kim_teaching_2021}, and for inferring and suppressing disturbances to dynamical systems \cite{skardal_detecting_2023,restrepo_suppressing_2023}.
Reservoir computers have two main advantages over other machine-learning architectures for these kinds of problems.
First, the training can be done relatively efficiently, since the output matrix can be found by a traditional least-squares optimization procedure.
Second, the reservoir can be any sufficiently complex nonlinear dynamical system.
This last property of reservoir computers has allowed the use of many physical systems as reservoirs, including field-programmable gate arrays \cite{kumar_efficient_2021}, reverberant cavity waves \cite{ma_short-wavelength_2022}, and even a bucket of water \cite{fernando_pattern_2003}.
Different physical implementations of reservoir computers offer advantages such as computation speed, low power, high dimensionality, or ease of implementation.
For a review of physical implementations of reservoir computers, see \cite{tanaka_recent_2019}.

Organic electrochemical transistors (OECTs) are devices potentially useful to interface electronic components with organic ones, which require little voltage and are relatively easy to manufacture.
Recently, there has been promising work demonstrating the potential for using OECTs as non-linear activation units for reservoir computers~\cite{cucchi_liquido_2023}.
In Ref.~\cite{pecqueur_neuromorphic_2018}, the authors used 12 disconnected OECTs as a reservoir to discriminate between square wave and triangle wave signals.
Ref.~\cite{cucchi_reservoir_2021} showed that small reservoirs built from networks of OECTs can accomplish prediction and classification tasks.
In Ref.~\cite{wang_organic_2023}, a reservoir computer consisting of an array of OECTs was integrated into a sensor platform for the measurement of ECG signals in the human body and used for real-time cardiac disease diagnosis.

Despite these encouraging works, there has not been a systematic exploration of the performance of OECT network reservoirs.
In this paper, we present a theoretical analysis of the use of networks of OECTs as reservoir computers.
Building on existing single-OECT mathematical models \cite{friedlein_microsecond_2016,bernards_steady-state_2007}, we demonstrate numerically how reservoirs of coupled OECTs are capable of successfully learning and replicating the dynamics of chaotic systems such as the Lorenz attractor.
We also explore how the prediction performance depends on the OECT operating parameters.

Our paper is organized as follows.
In Section \ref{sec:background}, we present the state-of-the-art in terms of OECT research, reservoir computing, and the two combined.
In Section~\ref{sec:model}, we present our theoretical model for both the transient response for an individual OECT and a network of OECTs and how to practically simulate these models.
In Section~\ref{sec:results}, we present our numerical results and comparison to existing reservoir computing methods.
Lastly, in Section~\ref{sec:discussion}, we discuss our results, their significance, and opportunities for future work.

\section{\label{sec:background} Background}

\subsection{\label{sec:background_rc} Reservoir computing}

Neural networks have gained popularity with the advent of large quantities of data and powerful computational resources and have created a revolution in machine learning for language models, climate, robotics, and many other applications.
These models can contain upwards of $10^{11}$ parameters \cite{brown_language_2020} and can be extremely computationally expensive.

Reservoir computers are a special case of a neural network first introduced in Refs.~\cite{jaeger_echo_2001,maass_real-time_2002} where an input signal is fed to a fixed, nonlinear system (the reservoir).
The internal state of the reservoir is then mapped to an output using a linear readout scheme that can then be trained to match a desired output signal.
This framework has two advantages: (i) the training occurs in the linear readout and is performed only once, which allows one to avoid expensive back-propagation schemes and use efficient ridge-regression algorithms, and (ii) since the internal structure of the reservoir is fixed, many types of nonlinear systems can be used as reservoirs, including recurrent neural networks \cite{gauthier_next_2021}, reverberant waves \cite{ma_short-wavelength_2022}, field-programmable gate arrays \cite{kumar_efficient_2021}, and various other neuromorphic systems \cite{tanaka_recent_2019,schuman_survey_2017,markovic_physics_2020}.
Our goal is to demonstrate a reservoir computer implementation using OECTs.

Before describing the OECT implementation of reservoir computers, we describe the general principles of reservoir computing when used to learn and predict the dynamics of a system based on a time series.
We assume that a trajectory of the dynamical system to be predicted is available in the interval $[-T,0]$ as $\{\boldu(-T + n\Delta t)\}_{n=0}^{T/\Delta T}$.
This time series can be used to force the dynamics of the reservoir, a high-dimensional dynamical system with state vector $\boldr(t)$.
As a result of the forcing, the state vector is updated as $\boldr(t+\Delta t) =\boldF(\boldr(t),\boldu(t))$, where the function $\boldF$ depends on the reservoir implementation.
The output from the reservoir, $\boldz(t)$, is typically constructed as a linear function of the internal states (although one could use a linear combination of more generally defined internal variables), so that
\begin{align}
\boldz(t) = \Wout\boldr(t),
\end{align}
where $\Wout$ is the output matrix.
For prediction, the entries of the output matrix are trained by minimizing the error between the actual output $\boldz(t)$ and the desired output,  $\boldu(t)$.
Thus, during training, the reservoir is trained so that an input $\boldu(t)$ gives an output close to $\boldu(t+\Delta t)$.
The minimization is typically done by ridge regression, i.e., by a least-squares minimization procedure that also includes a regularization term \cite{wyffels_stable_2008,cucchi_hands-reservoir_2022}, or with Lasso regression if sparseness is desired \cite{scardapane_distributed_2016}. 

After training, the reservoir can be operated in a closed-loop configuration, where the output at time $t$ is used as input for forecasting the output at time $t+\Delta t$.
In this autonomous operation mode the reservoir can, if successfully trained, produce a good approximation to the evolution of the dynamics for $t > 0$.
If the underlying dynamics are chaotic, the reservoir prediction will necessarily diverge away from the true evolution of the system after some time.
The length of time during which the approximation is good is referred to as the {\it Forecast Horizon} \cite{haluszczynski_good_2019}, and we use it as a measure of prediction quality.
We provide a precise definition of the forecast horizon in Sec.~\ref{sec:results}.

In the next sections, we will describe how a network of coupled OECTs can be used as the reservoir.
First, we present the mathematical description of a single OECT and then the scheme by which they are coupled on a network.

\begin{figure}[t]
    \centering
    \includegraphics[width=8.6cm]{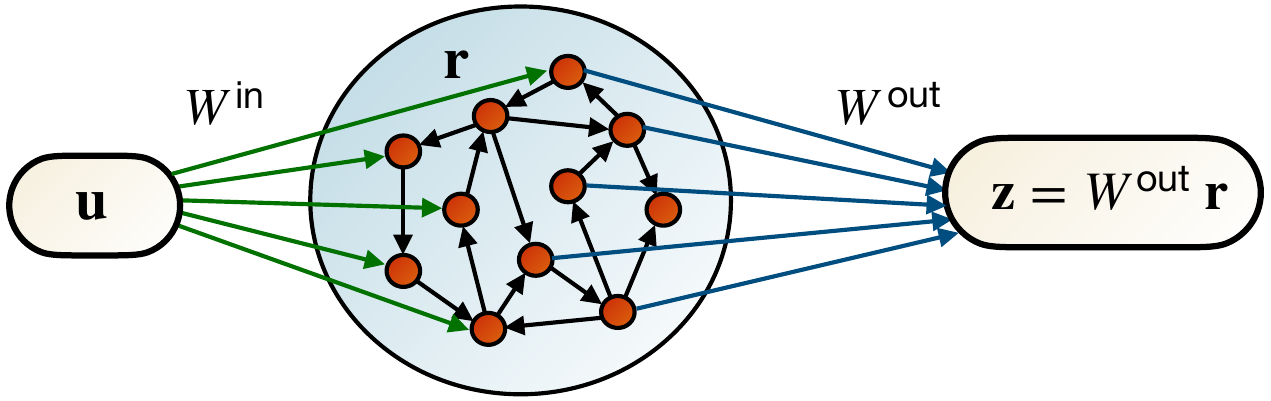}
    \caption{Implementation of a reservoir computer using a recurrent neural network.}
    \label{fig:cartoon}
\end{figure}

\subsection{\label{sec:background_oect} Single OECT model}

\begin{figure}
    \centering
    \includegraphics[width=8.4cm]{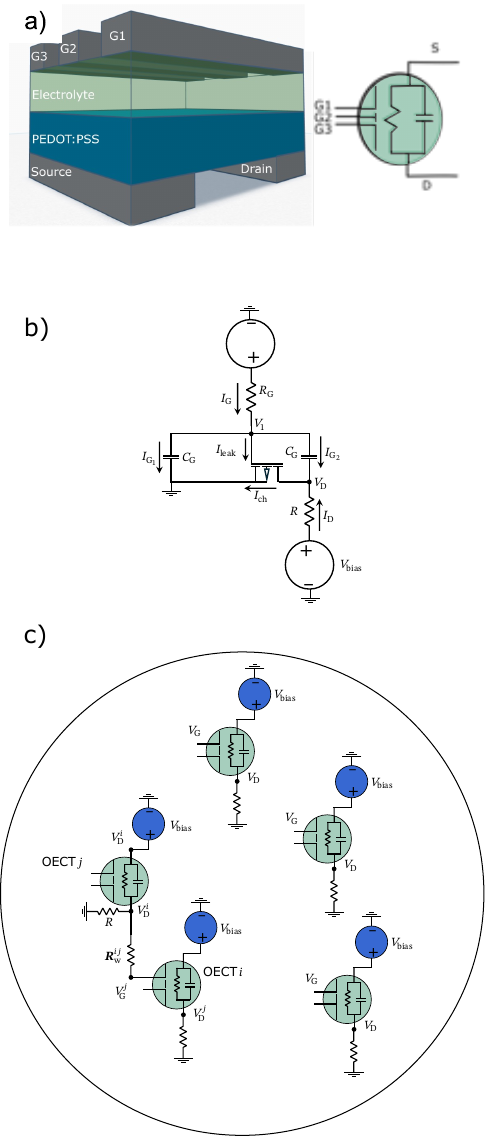}\\     
    \caption{a) A conceptual illustration of an example multi-gate OECT that serves as a node in the reservoir computer, in this case with three gate electrodes that receive signals from neighboring nodes. Also shown is the circuit symbol that is used in panel c). b) The circuit diagram of the device, showing the parameters used in the theoretical OECT model. c) The circuit diagram of an example OECT RC with a single representative connection between the drain of OECT $i$ and the gate of OECT $j$, weighted with resistor $\boldRW^{ij}$.}
        \label{fig:OECT_device_circuit}
\end{figure}

Organic electrochemical transistors (OECTs) are three-terminal organic electronic devices that rely on the transport of ions through an electrolyte to dope or de-dope a conductive polymer channel [see Fig.~\ref{fig:OECT_device_circuit}(a)].
The amount of doping or de-doping is regulated by the gate electrode voltage $\VG$, resulting in changes in the channel current $\Ich$ between the source and drain electrodes.
Due to their reliance on ion migration across a liquid-polymer interface and intercalations of ions into the conducting polymer channel, the degree of which depends on the history of voltages applied to the device, OECTs have highly nonlinear and history-dependent behavior.  This allows them to display neuromorphic behaviors \cite{tuchman_organic_2020} such as associative learning \cite{ji_mimicking_2021}.

Before considering reservoirs consisting of networks of coupled OECTs, we present a mathematical model governing the dynamics of a single OECT.
This model is based on previous models developed in Refs.~\cite{friedlein_microsecond_2016,perez_neuromorphic-based_2020}.
The circuit model of an OECT used here is shown in Fig.~\ref{fig:OECT_device_circuit}(b).
Of particular interest is the {\it gate voltage} $\VG$, which we will consider as the input variable to the OECT device, and the {\it drain voltage} $\VD$, which we will consider as the output variable.
Thus, given an input signal $\VG(t)$, our model specifies the output $\VD(t)$.

An important aspect of this implementation of OECTs into a reservoir computer is the resistor $R$ inserted between the drain electrode of the device and the applied voltage $\Vbias$.
This resistor is the basis for the connection weight between two devices, as the voltage generated at the gate electrode of the receiving device is determined by the voltage divider formed between $R$ and the drain-source channel of the sending device.
A unique aspect of this implementation is that the weight of the connection effectively changes according to the state of the sending device.
This dynamic nature of the connection weight is not present in typical neural networks, but it can be utilized to design useful circuits such as the neuromorphic-based Boolean circuits previously demonstrated in Ref.~\cite{perez_neuromorphic-based_2020}.

\begin{table}[b]
    \centering
    \begin{tabular}{ccl}
    \textbf{Notation}&\hspace{0.5cm} & \textbf{Definition}\\
        \hline\\
        $\VG$ && Gate voltage \\
        $V_1$ && Voltage at the ionic channel \\
        $\VD$ && Drain voltage \\
        $\Vbias$ && Offset voltage \\
        $\Vp$ && Pinch-off voltage \\
        $\RG$ && Effective gate resistance \\
        $R$ && Drain resistance \\
        $\CG$ && Effective gate capacitance \\
        $\Kp$ && Transconductance of the ionic channel\\
        $W$ && Width of the ionic channel \\
        $L$ && Length of the ionic channel
    \end{tabular}
    \caption{Table of relevant notation for the theoretical OECT model.}
    \label{tab:oect_notation}
\end{table}

The architecture and construction of neuromorphic OECTs are discussed in more detail in Section~\ref{sec:background_oect}, but here we present the relevant parameters of the model described in this section in Table~\ref{tab:oect_notation}.
Of all the quantities listed in Table~\ref{tab:oect_notation}, $\VG$, $\VD$, and $V_1$ are the only ones that vary in time.
The remainder are device-specific parameters and depend on the fabrication of each physical device.
In the approximation leading to the diagram in Fig.~\ref{fig:OECT_device_circuit}, the flow of ions in and out of the channel through the gate is approximated as an RC circuit with resistance $\RG$ and capacitance $\CG$.
The differential equation modeling the voltage at the channel can then be written as 
\begin{equation}
\frac{dV_1}{dt}=\frac{1}{\RG \CG}(\VG - V_1). \label{eq:v1_single_oect}
\end{equation}
From the gate voltage $\VG(t)$, Eq.~(\ref{eq:v1_single_oect}) determines $V_1(t)$. To determine the drain voltage $\VD$ from $V_1$, we note that, following \cite{friedlein_microsecond_2016}, these variables can be related using the following equations: 
\begin{align}
\IG&=\frac{\VG-V_1}{\RG},\\
\ID&=\Ich-\frac{\IG}{2},\\
\VD&=\Vbias -\ID R,\\
\Ich&=
\begin{cases}
\frac{-\Kp W}{2L} (V_1-\Vp )^2,& V_1-\VD > \Vp\\[0.05in]
0, & V_1 > \Vp, \VD\leq 0\\[0.05in]
-\frac{\Kp W}{L} \left(V_1-\Vp-\frac{\VD}{2}\right) \VD,& \text{else.}\end{cases}
\label{eq:single_oect}
\end{align}
The cases in the piecewise function correspond to the so-called saturation regime, cutoff regime, and linear regime, respectively \cite{bernards_steady-state_2007}.
Simplifying these equations, and solving for $\VD$, we obtain the following piecewise equation:
\begin{equation}
\VD=\begin{cases}
\left(
\begin{aligned}
&\Vbias+ a (\VG - V_1)\\
&+\frac{b}{2}(V_1 - \Vp)^2
&
\end{aligned}\right),
& V_1-\VD > \Vp\\[0.25in]
\Vbias + a (\VG - V_1), & V_1 > \Vp, \ \VD \leq 0\\[0.05in]
-\frac{1}{b}+(V_1 - \Vp) + \frac{\sqrt{\Delta}}{b}, & \text{else,}
\end{cases}
\label{eq:vd_single_oect}
\end{equation}
where
\begin{align*}
a &= \frac{R}{2\RG},\\
b &= \frac{\Kp W R}{L},\\
\Delta &= \max(2b \left[\Vbias + a(\VG - V_1)\right] \nonumber\\
&+ \left[b(V_1 - \Vp) - 1\right]^2, 0).
\end{align*}

In summary, given an input gate voltage $\VG(t)$, the output drain voltage $\VD(t)$ is found by solving the differential equation (\ref{eq:v1_single_oect}) for $V_1(t)$ and then calculating $\VD$ from Eq.~(\ref{eq:vd_single_oect}).

\section{\label{sec:model} OECT reservoir computer}

\subsection{Mathematical Model of OECT network}

In this section, we describe how a network of coupled OECTs can be used as a reservoir to predict the evolution of chaotic dynamical systems.
This theoretical implementation will be referred to as an OECT RC in the following.
We consider $N$ OECTs labeled $n = 1,2,\dots,N$, and indicate the variables for the $n$th OECT with a superscript $n$.
We construct a network of coupled OECTs (henceforth referred to as {\it nodes}), where node $m$ connects to node $n$ if the drain voltage of node $m$ connects to the gate voltage of node $n$ with resistance $\boldRW^{nm}$.
Using Kirchoff's laws, one can show that the gate voltage of node $n$ is given by
\begin{align}
\VG^n &= \frac{V_{1}^n}{\RG^n S^n} +  \sum_{m=1}^N \boldA^{nm}\VD^m,
\label{eq:vg_network}
\end{align}
where
\begin{align}
\boldA^{nm}&=\frac{1}{\boldRW^{nm}S^n},\label{eq:def_A}\\
S^n &= \frac{1}{\RG^n} + \sum_{m=1}^N \frac{1}{\boldRW^{nm}},\label{eq:def_S}
\end{align}
the superscript denotes the node index, and $\boldRW^{nm}\to \infty$ when node $m$ is not connected to node $n$.
Defining a vector $\boldf$ with entries $f^n = 1/(\RG^n S^n)$, we can express the gate voltage vector $\boldVG = [\VG^1, \VG^2,\dots,\VG^N]^T$ as $\boldVG = \boldA \boldVD + \boldf \otimes \boldV_1$, where $\otimes$ indicates elementwise multiplication.
For large enough values of $\RG$ and $N$, we can neglect the first terms in Eqs.~\eqref{eq:vg_network} and \eqref{eq:def_S}.

We note that when the first term in equation~\eqref{eq:vg_network} is neglected, the total gate voltage $\VG^n$ of a node can be expressed as a simple sum of output voltages generated by neighboring nodes in the network.
This implies that the concentration of ions transported to the conducting channel as a result of the contribution to $\VG^n$ from each neighbor is independent.
This assumption holds if the multiple inputs to the device are symmetric and provide the same ion transport distance through the electrolyte for each input.
Preliminary experimental results of multi-gate OECTs by the authors, not shown here, verified this to be a good assumption and achievable experimentally for at least a small number (3) of connected neighbors and for devices on the size scale of $\sim$1 mm.

\begin{table}[t]
    \centering
    \begin{tabular}{ccl}
    \textbf{Notation}&\hspace{0.5cm} & \textbf{Definition}\\
        \hline\\
        $N$ && Number of OECTs\\
        $\Win$ && The matrix encoding the input layer\\
        $\Wout$ && The matrix encoding the output layer\\
        $\boldRW$ && Matrix of weighting resistors\\
        $\boldA$ && The effective adjacency matrix
    \end{tabular}
    \caption{Table of relevant notation for the theoretical network OECT model.}
    \label{tab:network_notation}
\end{table}

The evolution of the reservoir without external input is given by the coupled equations
\begin{align}
&\frac{dV_1^n}{dt}=\frac{1}{\RG^n \CG^n}(\VG^n-V_1^n), \label{eq:v1_oect_network}\\
&\VG^n = f^n V_{1}^n +  \sum_{m=1}^N \boldA^{nm} \VD^m,\label{eq:vg_oect_network}\\
&\VD^n =\begin{cases}
\left(
\begin{aligned}
&\Vbias^{n}+a^n (\VG^n - V_1^n)\\[0.05in]
&+\frac{b^n}{2}(V_1^n - \Vp^n)^2\\
\end{aligned}\right),
& V_1^n-\VD^n > \Vp^n\\[0.25in]
\Vbias^{n} + a^n (\VG^n - V_1^n), & V_1^n > \Vp^n, \ \VD^n\leq 0\\[0.05in]
-\frac{1}{b^n}+(V_1^n-\Vp^n) + \frac{\sqrt{\Delta^n}}{b^n}, & \text{else,}
\end{cases}\label{eq:vd_oect_network}
\end{align}
where
\begin{align*}
a^n &= \frac{R^n}{2\RG^n},\\
b^n &= \frac{\Kp^n W^n R^n}{L^n},\\
\Delta &= \max(2b^n \left[\Vbias^n + a^n(\VG^n - V_1^n)\right] \nonumber\\
&+ \left[b^n(V_1^n - \Vp^n) - 1\right]^2, 0),
\end{align*}
and $n$ superscripts are indices.

\subsection{Training and Prediction}

Having described the dynamics of the network of OECTs (the reservoir), we now discuss how it can be used for the prediction of a nonlinear dynamical system.
As discussed in Sec.~\ref{sec:background_rc}, the reservoir takes as input $\boldu(t) \in \mathbb{R}^D$ during its training phase.
We assume that the input is fed to the gate voltage of the reservoir nodes via an input matrix $\Win$.
The resulting equations are given by
\begin{align}
&\frac{dV_1^n}{dt}=\frac{1}{\RG^n \CG^n}(\VG^n-V_1^n), \label{eq:v1_oect_network_training}\\
&\VG^n = f^n V_{1}^n +  \sum_{m=1}^N \boldA^{nm} \VD^m + \sum_{k=1}^D \Win^{nk} u_{k}(t),\label{eq:vg_oect_network_training}\\
&\VD^n =\begin{cases}
\left(
\begin{aligned}
&\Vbias^n + a^n (\VG^n - V_1^n)\\[0.05in]
&+\frac{a^n}{2}(V_1^n-\Vp^n )^2
\end{aligned}\right),
& V_1^n-\VD^n > \Vp^n\\[0.25in]
\Vbias^n + a^n (\VG^n - V_1^n), & V_1^n > \Vp^n, \ \VD^n\leq 0\\[0.05in]
-\frac{1}{b^n}+(V_1^n-\Vp^n) + \frac{\sqrt{\Delta^n}}{b^n}, & \text{else.}
\end{cases}
\label{eq:vd_oect_network_training}
\end{align}
Again, the $n$ superscripts indicate the indices of each parameter.
In this paper, we will focus on predicting the dynamics of the chaotic Lorenz system,
\begin{align}
\frac{dx}{dt} &= \sigma (y - x),\label{eq:lorenz1}\\
\frac{dy}{dt} &= x(\rho - z) - y,\label{eq:lorenz2}\\
\frac{dt}{dt} &= x y - \beta z,\label{eq:lorenz3}
\end{align}
with $\sigma = 10$, $\rho = 28$, and $\beta = 8/3$, where the state of the system is given by the vector $\boldu = [x, y, z]^{\text{T}}$.

\subsubsection{Training}

In the following, we assume that we have available a time series of the state vector obtained from Eqs.~\eqref{eq:lorenz1}-\eqref{eq:lorenz3}, $\{\boldu(-T),\boldu(-T+\Delta t)\dots,\boldu(-\Delta t), \boldu(0)\}$, for time $t \in [-T,0)$, that can be used to train the reservoir to predict the values of $\boldu(t)$ for $t >0$.
The discrete time series could be obtained from a numerical solution Eqs.~\eqref{eq:lorenz1}-\eqref{eq:lorenz3} or, in more general cases, from discrete-time measurements of an experimental system.

To train the reservoir, we integrate numerically Eqs.~\eqref{eq:v1_oect_network_training}-\eqref{eq:vd_oect_network_training} using the same step $\Delta t$, and record the time-series of drain voltages $\{\boldVD(-T + j \Delta t)\}_{j = 0}^{T/\Delta t}$,  where $\boldVD = [\VD^1, \VD^2,\dots,\VD^N]^{\text T}$.
[In the context of Sec.~\ref{sec:background_rc}, the drain voltage vector $\boldVD$ corresponds to the internal state of the reservoir $\boldr(t)$.]
The output matrix $\Wout$ is then chosen to minimize
\begin{align}
E = &\sum_{j = 0}^{T/\Delta t} \| \Wout \boldVD(-T+j\Delta t) - \boldu(-T+j\Delta t) \|_2^2\nonumber \\
&+\alpha \|\Wout\|_2^2,
\end{align}
where the first term attempts to choose $\Wout$ so that the reservoir output $\boldz = \Wout \boldVD$ is as close as possible to the predicted state of the system $\boldu$, and the second term penalizes large coefficients in $\Wout$ to avoid over-fitting.

We assumed that the time discretization $\Delta t$ is small enough for accurate numerical solution of {\it both} Eqs.~(\ref{eq:v1_oect_network_training})-(\ref{eq:vd_oect_network_training}) and (\ref{eq:lorenz1})-(\ref{eq:lorenz3}). If, for example, a smaller time step $\delta t < \Delta t$ is required to numerically integrate the reservoir equations (\ref{eq:v1_oect_network_training})-(\ref{eq:vd_oect_network_training}), one can use the same procedure by setting $u(j\Delta t) = u(-T + j\lfloor \Delta t / \delta t \rfloor \delta t)$ in the numerical integration.

\subsubsection{Prediction}

Once the output matrix is chosen, the reservoir is trained to predict the value of $\boldu(t+\Delta t)$ when previous values of $\boldu(t)$ have been used to force it.
Therefore, one can take the predicted value of $\boldu$ at time $t + \Delta t$, $\boldz(t)$, and use it as input to predict the value of $\boldu$ at time $t + 2\Delta t$ and continue in this fashion to obtain a time series of predictions.
In this autonomous, or {\it closed-loop} mode, the evolution of the reservoir is given by  
\begin{align}
&\frac{dV_1^n}{dt}=\frac{1}{\RG^n \CG^n}(\VG^n-V_1^n), \label{eq:v1_oect_network_prediction}\\
&\VG^n = f^n V_{1}^n +  \sum_{m=1}^N \boldA^{nm} \VD^m + \sum_{k=1}^D \Win^{nk} \sum_{m = 1} \Wout^{km} \VD^m(t),\label{eq:vg_oect_network_prediction}\\
&\VD^n =
\begin{cases}
\left(
\begin{aligned}
&\Vbias+a^n (\VG^n - V_1^n)\\[0.05in]
&+\frac{b^n}{2}(V_1^n-\Vp^n )^2
\end{aligned}\right),
& V_1^n-\VD^n > \Vp^n\\[0.25in]
\Vbias^n + a^n (\VG^n - V_1^n), & V_1^n > \Vp^n, \ \VD^n\leq 0\\[0.05in]
-\frac{1}{b^n}+(V_1^n-\Vp^n) + \frac{\sqrt{\Delta^n}}{b^n}, & \text{else,}
\end{cases}
\label{eq:vd_oect_network_prediction}
\end{align}
which is solved with the same numerical scheme used for training.
Again, the $n$ superscript is an index.
Then, the prediction of the reservoir for the future evolution of $\boldu$ is the time series $\{\Wout \boldVD(i\Delta t)\}$, for $i = 1,2,\dots$.

\section{\label{sec:results} Results}

In this section, we present our numerical results for our theoretical model of an OECT RC.
We quantify the quality of our results via the \textit{Forecast Horizon} (FH) (e.g., see Ref.~\cite{haluszczynski_good_2019}), which quantifies the time during which the predictions of the reservoir, $\hat{\boldu}$, for the actual state of a system, $\boldu$, remain good. The Forecast Horizon can be defined as
\begin{align}
\text{FH}(\delta) = \argmin_{t}\{ \Vert \boldu(t) - \hat{\boldu}(t) \Vert > \delta \mid t > 0\},
\end{align}
where $\Vert \cdot \Vert$ is the $L_2$ norm.
The Forecast Horizon describes the minimum time at which the difference between the predicted and actual trajectories exceeds a specified error tolerance, $\delta$. In the following, we use $\delta=5$.

We construct OECT reservoir computers by first simulating a distribution of operating characteristics for OECTs and, second, by defining random strengths of connection between them.
When fabricating OECTs, there is natural variation in their operating characteristics \cite{gentile_mathematical_2020}.
In addition, variable properties can be helpful in capturing different timescales of an input signal \cite{pecqueur_neuromorphic_2018,gentile_mathematical_2020}.
In our case, we draw all parameters from a gamma distribution to capture variation in the device fabrication. 
In cases where the standard deviation is zero, we draw the parameters from a delta distribution centered at the mean. 
These parameters ($\Vbias$, $\Vp$, $W$, and $L$) represent device characteristics over which we have greater control in the fabrication process.
The mean and standard deviation values chosen for each operating parameter are tabulated in Table~\ref{tab:operating_characteristics}.
\begin{table}[b]
    \centering
    \begin{tabular}{cclclcl}
    \textbf{Quantity}&\hspace{0.5cm} & \textbf{Mean}&\hspace{0.5cm} & \textbf{Std. Dev.}&\hspace{0.5cm} & \textbf{Units}\\
        \hline
        $\Vbias$ && $-0.5$ && $0$ && Volt \\
        $\Vp$ && $-0.6$ && $0$ && Volt \\
        $R$ && $500$ && $100$ && Ohm \\
        $\RG$ && $2.7 \times 10^{4}$ && $2.7 \times 10^{3}$ && Ohm \\
        $\CG$ && $8.98 \times 10^{-7}$ && $8.98 \times 10^{-8}$ && Farad \\
        $\Kp$ && $5.82 \times 10^{-4}$ && $5.82 \times 10^{-5}$ && Siemens\\
        $W$ && $1.01\times 10^{-4}$ && 0 && Meter\\
        $L$ && $2.0 \times 10^{-4}$ && 0 && Meter\\
    \end{tabular}
    \caption{Table specifying the theoretical OECT operating parameters. Each parameter is drawn from a gamma distribution specified by its mean and standard deviation.}
    \label{tab:operating_characteristics}
\end{table}

\begin{figure}
    \centering
    \includegraphics[width=8.6cm]{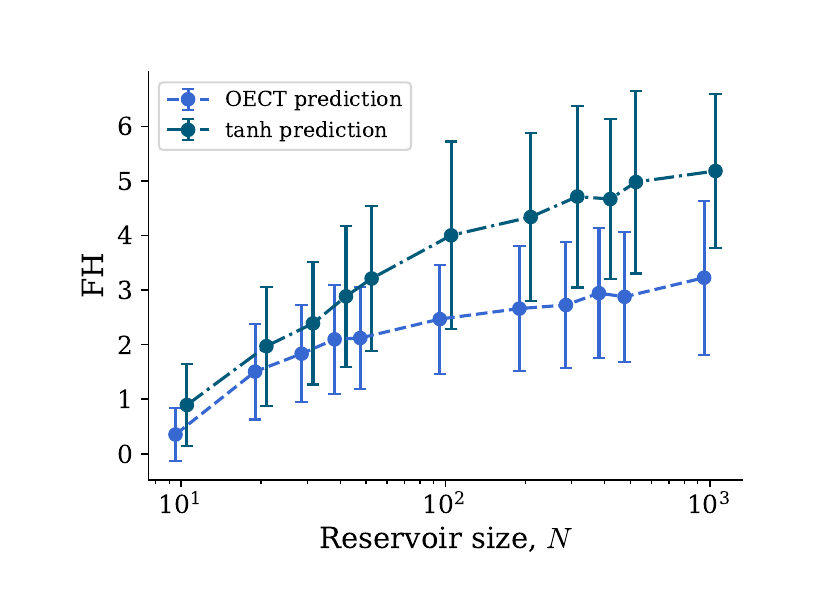}
    \caption{\label{fig:FH_vs_n} A plot of the forecast horizon of the Lorenz attractor described in Eqs.~\eqref{eq:lorenz1}-\eqref{eq:lorenz3} with respect to the reservoir size, $N$, for (1) an RC constructed according to Ref.~\cite{pathak_using_2017} (teal dash-dotted line) and (2) an OECT RC (blue dashed line). The prediction task is repeated $10^2$ times for each reservoir size. The connected lines represent the mean forecast horizon and the error bars represent the standard deviation in the forecast horizon.}
\end{figure}

The input layer $\Win$ and the reservoir connection matrix $\boldA$ were constructed as follows.
The input layer, which distributes components of the signal to each OECT in the reservoir, is a random $3 \times N$ matrix, with random entries drawn from the distribution $\Win^{nm}\sim \text{Uniform}(-\sigma, \sigma)$, where $\sigma=10^{-3}$.
For all numerical experiments, $N=100$, except when varying the reservoir size, as in Fig.~\ref{fig:FH_vs_n}.
The matrix $\boldA$ determining the connections between OECTs is specified by (1) the existence and non-existence of links between OECTs and (2) the strength of these connections. We use the directed \ER network model without self-loops to generate random directed networks. The links created in this process are weighted by first sampling resistance values from the distribution $\boldRW^{nm} \sim \text{Uniform}(100, 500)$ and then constructing the effective adjacency matrix $\boldA$ from Eq.~\eqref{eq:def_A} and \eqref{eq:def_S}. As mentioned before, large enough $\RG$ values allow us to neglect the $1/\RG^n$ term in Eq.~\eqref{eq:def_S}. Without this term, the matrix $A$ in Eq~\eqref{eq:def_A} is row-stochastic. Therefore, for large values of $\RG$ the  spectral radius of $A$ is  $\rho(\boldA) \approx 1$, which is close to the optimal value determined in Ref.~\cite{lu_reservoir_2017}.
Similarly, assuming large values of $\RG$, we neglect the $V_1^n/(\RG^n S^n)$ term in Eq.~\eqref{eq:vg_network}.
When training the output layer, we chose a ridge regression parameter of $\alpha=10^{-7}$ for all numerical experiments except when varying the ridge regression parameter as shown in Fig.~\ref{fig:FH_vs_alpha}.
The dynamical system of interest is the Lorenz attractor described in Eqs.~\eqref{eq:lorenz1}-\eqref{eq:lorenz3} with an initial condition of $\boldu_0 = \boldu(10)$, where $\boldu(t)$ is the solution to Eqs.~\eqref{eq:lorenz1}-\eqref{eq:lorenz3} with $\boldu(0) = [-7.4, -11.1, 20] + \boldsymbol{\delta}$, where $\boldsymbol{\delta}_i \sim \text{Normal}(0, \, 0.1)$. This is done so that (1) there is variation in the set of initial conditions and (2) we run the dynamical system for sufficient time so that the trajectory relaxes to the attractor manifold. We sample 100 initial conditions and 100 corresponding reservoir architectures for each parameter value of interest.

\begin{figure}
    \centering
    \includegraphics[width=8.6cm]{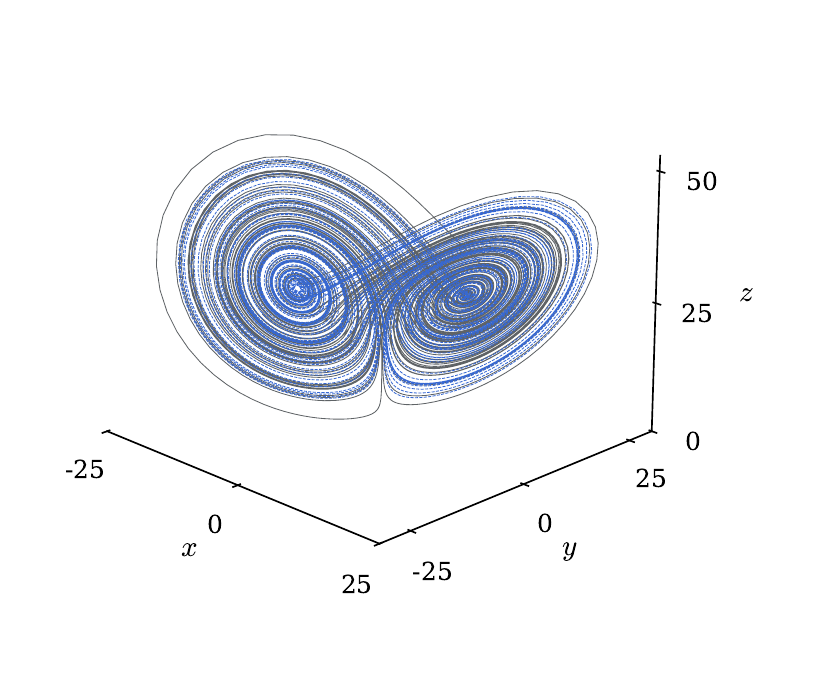}
    \caption{\label{fig:3d} A 3D plot of the Lorenz attractor (grey lines) and the prediction of the Lorenz attractor using an OECT RC (blue lines) of size $N=100$.}
\end{figure}

\begin{figure}
    \centering
    \includegraphics[width=8.6cm]{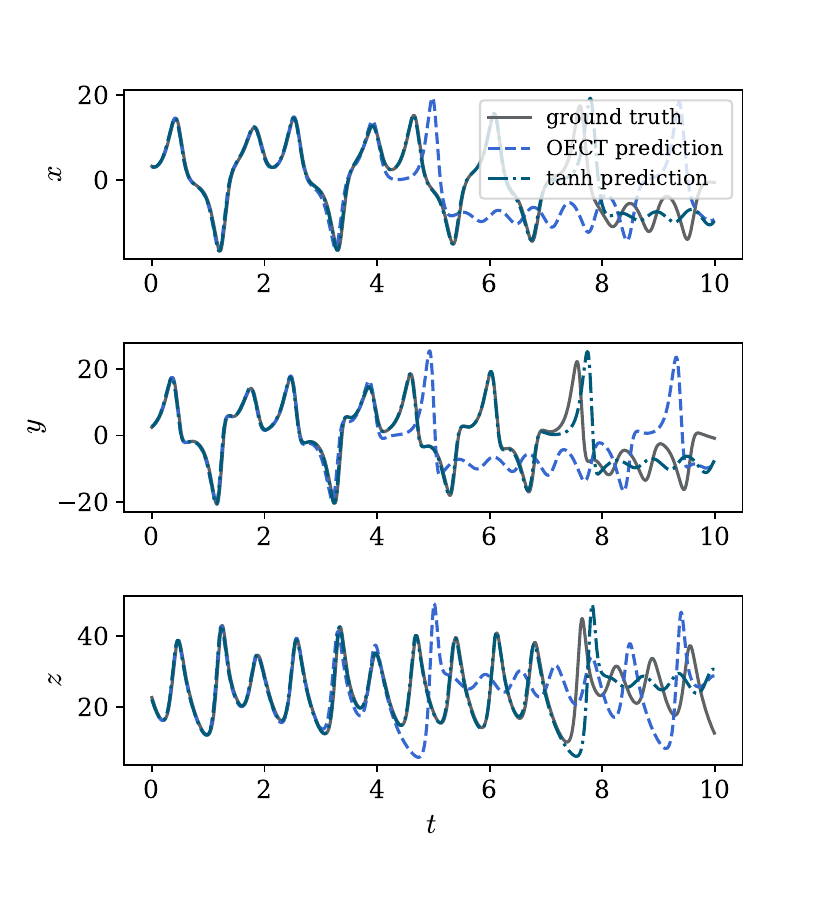}
    \caption{\label{fig:combined_time_series} A plot of the $x$, $y$, and $z$ components of the Lorenz attractor for the true time series (solid grey line), the time series predicted using the RC implementation described in Ref.~\cite{pathak_using_2017} (teal dashed-dotted line), and the time series predicted using the theoretical model of an OECT RC (blue dashed line). In both cases, $N = 100$.}
\end{figure}

In Fig.~\ref{fig:3d} we plot the Lorenz attractor obtained from numerical solution of Eqs.~(\ref{eq:lorenz1})-(\ref{eq:lorenz3}) (grey) and the predicted attractor from the OECT RC (blue). The OECT reservoir computer captures well the climate of the Lorenz attractor. In Fig.~\ref{fig:combined_time_series} we plot the $x$, $y$, and $z$ variables from the Lorenz system obtained from numerical solution of Eqs.~(\ref{eq:lorenz1})-(\ref{eq:lorenz3}), which we refer to as ``ground truth'' (solid black lines), the prediction from the OECT reservoir computer (dashed blue lines), and the prediction from a standard reservoir computer implementation with a hyperbolic tangent activation function (e.g., see \cite{pathak_using_2017}) (dotted-dashed blue lines). We see that a small reservoir computer can accurately forecast the dynamics for a short time with comparable performance to standard reservoir computer implementations. These figures illustrate proof of concept; OECTs can be used to construct a reservoir computer with predictive capabilities. Because OECTs are physical devices, it is useful to understand how their operating parameters and connections between them affect their performance on prediction tasks.

We explore several basic properties of OECT RCs, specifically the reservoir size, the pinch-off voltage, and the density of connections. These properties are by no means exhaustive, but provide a useful starting point for future exploration.

In Fig.~\ref{fig:FH_vs_n} we plot the FH obtained from the OECT reservoir computer (blue dashed line) and from a reservoir computer implementation using  hyperbolic tangent activation functions as in \cite{pathak_using_2017} (dotted-dashed green line) as a function of the reservoir size $N$. We observe two things: first, the performance of the OECT RC performs, on average, similarly to the standard implementation of the reservoir computer for small reservoir sizes and second, the variation in performance is similar for both the OECT RC and the standard RC.

\begin{figure}
    \centering
    \includegraphics[width=8.6cm]{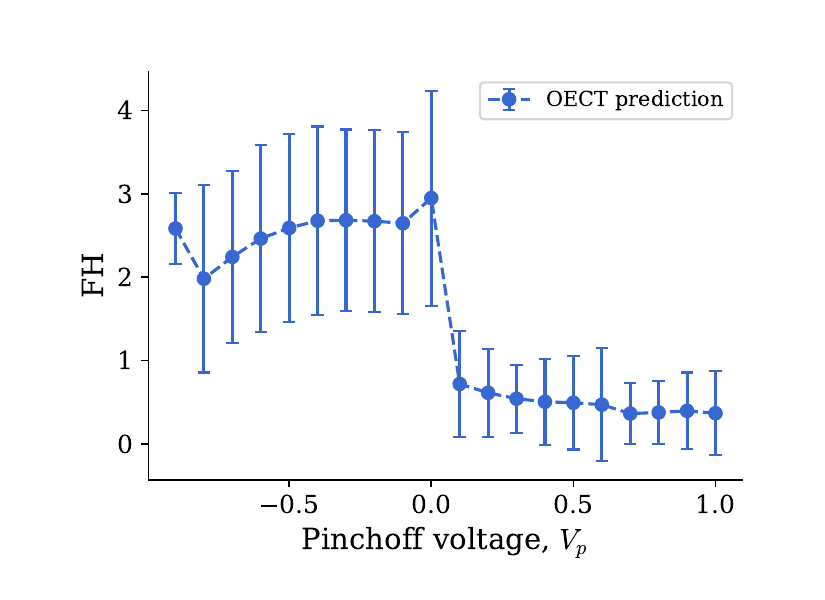}
    \caption{\label{fig:FH_vs_pinchoff} A plot of the forecast horizon of the Lorenz attractor described in Eqs.~\eqref{eq:lorenz1}-\eqref{eq:lorenz3} with respect to the pinch-off voltage, $\Vp$ for an OECT RC (blue dashed line) of size $N=100$. The prediction task is repeated $10^2$ times for each value of $\Vp$. The connected lines represent the mean forecast horizon and the error bars represent the standard deviation in the forecast horizon.}
\end{figure}
We have found that the pinch-off voltage $\Vp$ has a large effect on the performance of the OECT RC. In Fig.~\ref{fig:FH_vs_pinchoff} we plot the FH obtained from the OECT RC versus the pinch-off voltage $\Vp$. We see that the pinch-off voltage strongly influences the length of time that the OECT RC can predict into the future.
This is because Eq.~\eqref{eq:vd_single_oect} specifies three operating regimes: the linear regime, the cutoff regime, and the saturation regime.
The pinch off voltage $\Vp$ defines the boundary between the linear and saturation regimes \cite{ohayon_guide_2023}. Reducing the value of $\Vp$ shifts the typical operating point of the devices into the nonlinear saturation regime.
This nonlinearity gives the OECT RC more predictive power than would be expected if the devices remained in the linear regime.
Recent work has highlighted the importance of shifting the activation functions of nodes in a reservoir computer such that they are operating more frequently in the nonlinear regime \cite{hurley_tuning_2023}.
Fig.~\ref{fig:FH_vs_pinchoff} shows the improvement in FH with decreasing values of $\Vp$, continuing through zero to negative values. We note that variation of $\Vp$ across a range of both negative and positive values can be implemented experimentally via an ambipolar OECT \cite{stein_ambipolar_2022}.

\begin{figure}
    \centering
    \includegraphics[width=8.6cm]{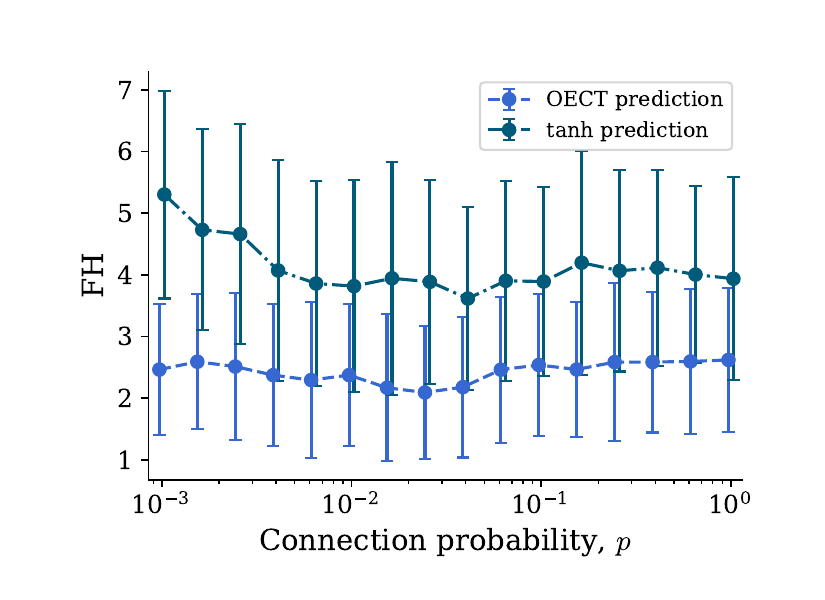}
    \caption{\label{fig:FH_vs_p} A plot of the forecast horizon of the Lorenz attractor described in Eqs.~\eqref{eq:lorenz1}-\eqref{eq:lorenz3} with respect to the connection probability, $p$, for (1) an RC constructed according to Ref.~\cite{pathak_using_2017} (teal dash-dotted line) and (2) an OECT RC (blue dashed line). The prediction task is repeated $10^2$ times for each value of $p$. The connected lines represent the mean forecast horizon and the error bars represent the standard deviation in the forecast horizon. In both cases, $N = 100$.}
\end{figure}

Lastly, a characteristic of the OECT RC that is much easier to change than the fabrication of physical devices is how the devices are connected to each other. In order to study how the average number of connections to and from each OECT, parameterized by the connection probability $p$, affects the reservoir performance, we plot the FH as a function of $p$ in Fig.~\ref{fig:FH_vs_p}.
We see that there is no strong dependence of the FH on the connection probability $p$. This corroborates the findings in Ref.~\cite{lu_reservoir_2017}, where most network densities produce the same amount of error.

\section{\label{sec:discussion} Discussion}

In this paper, we have presented a theoretical framework as a proof of concept that reservoir computers implemented with OECTs as the nonlinear activation units can be an effective way to predict complex dynamics. We have examined the effect of different operating parameters on the predictive power of these systems and have managed to recreate the dynamics of the Lorenz attractor with a reservoir computer running autonomously.

This theoretical framework excludes phenomena present in experimental implementations.
Our model assumes no resistive losses in the network of OECTs and assumes that the voltages provided to the gate of an OECT are linearly additive.
When implementing such a reservoir computer, it could prove necessary to include these features in our theoretical model to accurately model these systems.
This could, in fact, be advantageous as it would further increase the complexity of the reservoir dynamics.

This proof of concept will be useful in the design and fabrication of physical devices and can provide a theoretical framework to accompany the operation of physical devices.
Future work can more comprehensively explore the role that network structure and device parameters have on the performance of these reservoir computers.
The optimal OECT operating parameters and reservoir architecture will also likely depend on the characteristic time-scale of the dynamics of interest and this should be further explored.

\section*{Data availability} All code and data used in this study are available on \href{https://github.com/Signed-B/neuromorphic-computing-on-OECTs}{Github} and at Ref.~\cite{hyde_code_2024}.

\appendix

\section{Examining the effect of the ridge regression parameter}

The ridge regression parameter, $\alpha$, serves as an $L_2$ penalty on the magnitude of the coefficients in the output layer. As $\alpha \to 0$, the solution approaches that of the least-squares problem. The prediction accuracy of conventional reservoir computers is sensitive to $\alpha$ \cite{hurley_tuning_2023} and we explore whether this is also the case for OECT RCs.

\begin{figure}
    \centering
    \includegraphics[width=8.6cm]{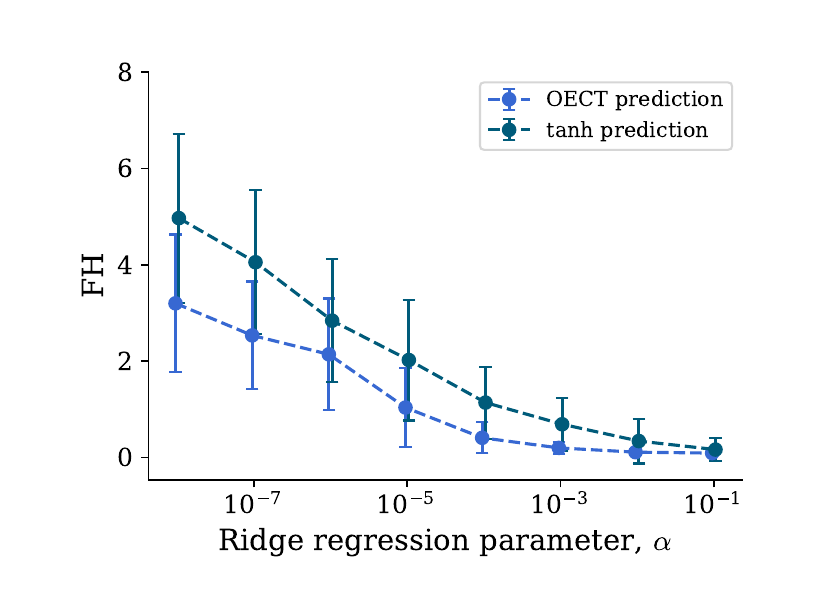}
    \caption{\label{fig:FH_vs_alpha} A plot of the forecast horizon of the Lorenz attractor described in Eqs.~\eqref{eq:lorenz1}-\eqref{eq:lorenz3} with respect to the ridge regression parameter, $\alpha$, for (1) an RC constructed according to Ref.~\cite{pathak_using_2017} (teal dash-dotted line) and (2) an OECT RC (blue dashed line). The prediction task is repeated $10^2$ times for each value of $\alpha$. The connected lines represent the mean forecast horizon and the error bars represent the standard deviation in the forecast horizon. In both cases, $N = 100$.}
\end{figure}

In Fig.~\ref{fig:FH_vs_alpha}, we see that the forecast horizon is increased with decreasing values of $\alpha$. Based on this plot, we chose a value of $\alpha = 1 \times 10^{-7}$ for all results presented in this paper.

\bibliography{references}

\end{document}